\documentclass[10pt,twocolumn,letterpaper]{article}

\usepackage[pagenumbers]{cvpr} 

\usepackage{colortbl}
\usepackage{multirow}
\usepackage{bbding}
\usepackage{algorithm}
\usepackage{algorithmic}

\definecolor{cvprblue}{rgb}{0.21,0.49,0.74}
\usepackage[pagebackref,breaklinks,colorlinks,allcolors=cvprblue]{hyperref}

\def\paperID{22670} 
\def\confName{CVPR}
\def\confYear{2026}

\title{Disentangle Object and Non-object Infrared Features via Language Guidance}

\author{
Fan Liu\textsuperscript{†}, 
Ting Wu, 
Chuanyi Zhang, 
Liang Yao, 
Xing Ma,
Yuhui Zheng
\\\\
Hohai University
\\
{\tt\small \textsuperscript{†}Corresponding Author
}
\\
{\tt\small Email: fanliu@hhu.edu.cn}
}

\begin{document}
\maketitle
\begin{abstract}
Infrared object detection focuses on identifying and locating objects in complex environments (\eg, dark, snow, and rain) where visible imaging cameras are disabled by poor illumination. However, due to low contrast and weak edge information in infrared images, it is challenging to extract discriminative object features for robust detection. To deal with this issue, we propose a novel vision-language representation learning paradigm for infrared object detection. An additional textual supervision with rich semantic information is explored to guide the disentanglement of object and non-object features. Specifically, we propose a Semantic Feature Alignment (SFA) module to align the object features with the corresponding text features. Furthermore, we develop an Object Feature Disentanglement (OFD) module that disentangles text-aligned object features and non-object features by minimizing their correlation. Finally, the disentangled object features are entered into the detection head. In this manner, the detection performance can be remarkably enhanced via more discriminative and less noisy features. Extensive experimental results demonstrate that our approach achieves superior performance on two benchmarks: M\textsuperscript{3}FD (83.7\% mAP), FLIR (86.1\% mAP). Our code will be publicly available once the paper is accepted.
\end{abstract}
    
\section{Introduction}
\label{sec:intro}
As a fundamental task in computer vision, object detection aims to identify and locate objects in images and videos. Unlike visible images, infrared ones are formed by thermal radiation, which can properly work under poor illumination. This advantage over RGB data allows infrared object detection (IROD) to have extensive applications in multiple scenarios, such as military reconnaissance~\cite{military}, autonomous driving~\cite{driving}, and security surveillance~\cite{monitoring}.

Since the image-forming principles of visible and infrared images are different, they can provide complementary information. Therefore, a large number of works fuse visible and infrared images for detection~\cite{DEYOLO-RGB-IR-fusion1,CrossFormer-RGB-IR-fusion2,RGB-IR-fusion3,Re-identification}. Nevertheless, obtaining precisely paired visible and infrared data has a high cost due to expensive dual-modality cameras and manual image matching. This drawback makes the infrared-visible fusion paradigms less practical.
\begin{figure}[t]
    \centering
    \includegraphics[width=0.99\linewidth]{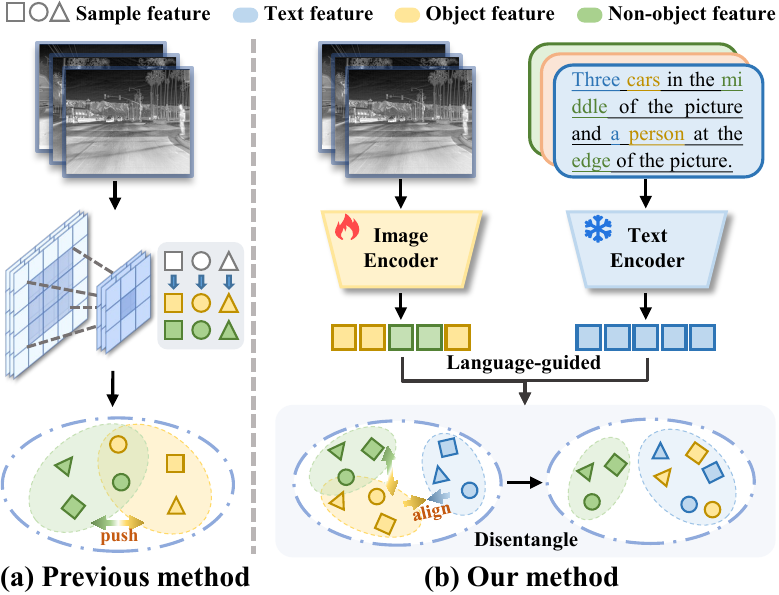}
    \caption{
    Comparison between previous method and our LGFD for feature disentanglement.
    (a) The previous method typically struggle to extract object features through convolution in an entangle feature space. (b) Our LGFD adopts a language-guided representation learning strategy. It disentangles the object and non-object feature via textual semantic information guidance.}   
    \label{fig1}
\end{figure}

Another branch of IROD research is merely utilizing infrared images without paired visible data. 
However, learning discriminative visual features from infrared images tends to be more challenging compared to learning from visible modality. Specifically, visible images have rich color and texture information, while infrared images primarily reflect the thermal radiation information of the object's surface rather than its visual appearance.
Consequently, infrared images typically have low contrast and weak edge information due to inapparent texture and colorlessness.
This trait makes it difficult to distinguish objects from the background in infrared images. From this observation, it is intuitive to disentangle the object and background features to enhance the detection performance. 
The previous method~\cite{YOLO-CIR-IRfinetune1} attempts to design feature extraction modules to separate object and background features. 
Nevertheless, merely relying on simple convolution modules to separate object features from global features without reliable supervision tends to result in an insufficient feature separation.
Moreover, the extracted features of the object might lack interpretability. 

Motivated by the previous object feature learning approach~\cite{YOLO-CIR-IRfinetune1} (Figure~\ref{fig1} (a)), we aim to disentangle object and non-object features in an effective and interpretable manner. From the perspective of interpretability, utilizing image text descriptions with object categories and location information can provide semantic information of objects. The recent success of vision-language research~\cite{CLIP,BLIP,ALIGN} has proven the potential of text modality in representation learning.
With the assistance of language, we can properly obtain object features via aligning visual and textual features. 

To this end, we propose a \textbf{Language-guided Feature Disentanglement (LGFD)} approach to separate object and non-object features via the guidance of textual semantic information, as illustrated in Figure~\ref{fig1} (b).
Through aligning text and object features, our LGFD can achieve more remarkable object and non-object feature disentanglement over the previous approach~\cite{YOLO-CIR-IRfinetune1}.
Specifically, we divide visual features channels into object and non-object groups. Then, we propose a Semantic Feature Alignment (SFA) module to align the object features with the corresponding text features. In this manner, we transform the conventional visual detection pipeline into a vision-language representation learning paradigm, enabling the detection model to learn semantic information from texts.
The captions paired with the images are derived from the sample annotations and provide critical object semantic information.Moreover, we design an Object Feature Disentanglement (OFD) module that disentangles text-aligned object features and non-object features by minimizing the correlation between them. This essential disentanglement allows the model to focus on salient features relevant to the task while effectively reducing the influence of irrelevant non-object information. Finally, the disentangled object features are entered into the detection head. With noisy non-object information suppressed, the detection performance can be significantly boosted. 

Our contributions are summarized as follows:
\begin{itemize}
    \item We propose a novel vision-language representation learning paradigm that leverages textual semantic information to guide infrared object detection. To the best of our knowledge, \textbf{we are the first to investigate infrared image-text exploitation for detection tasks.}

    \item We develop an interpretable Language-guided Feature Disentanglement framework. The proposed Semantic Feature Alignment (SFA) module learns object features from textual semantic information while the Object Feature Disentanglement (OFD) module enhances object features via decoupling non-object ones.

    \item We conduct comprehensive experiments on four infrared datasets to validate our method. Experimental results indicate that our approach surpasses all existing infrared object detection approaches and even outperforms many RGB-IR fusion detection frameworks. 
\end{itemize}

\begin{figure*}[t]
    \centering
    \includegraphics[width=0.99\linewidth]{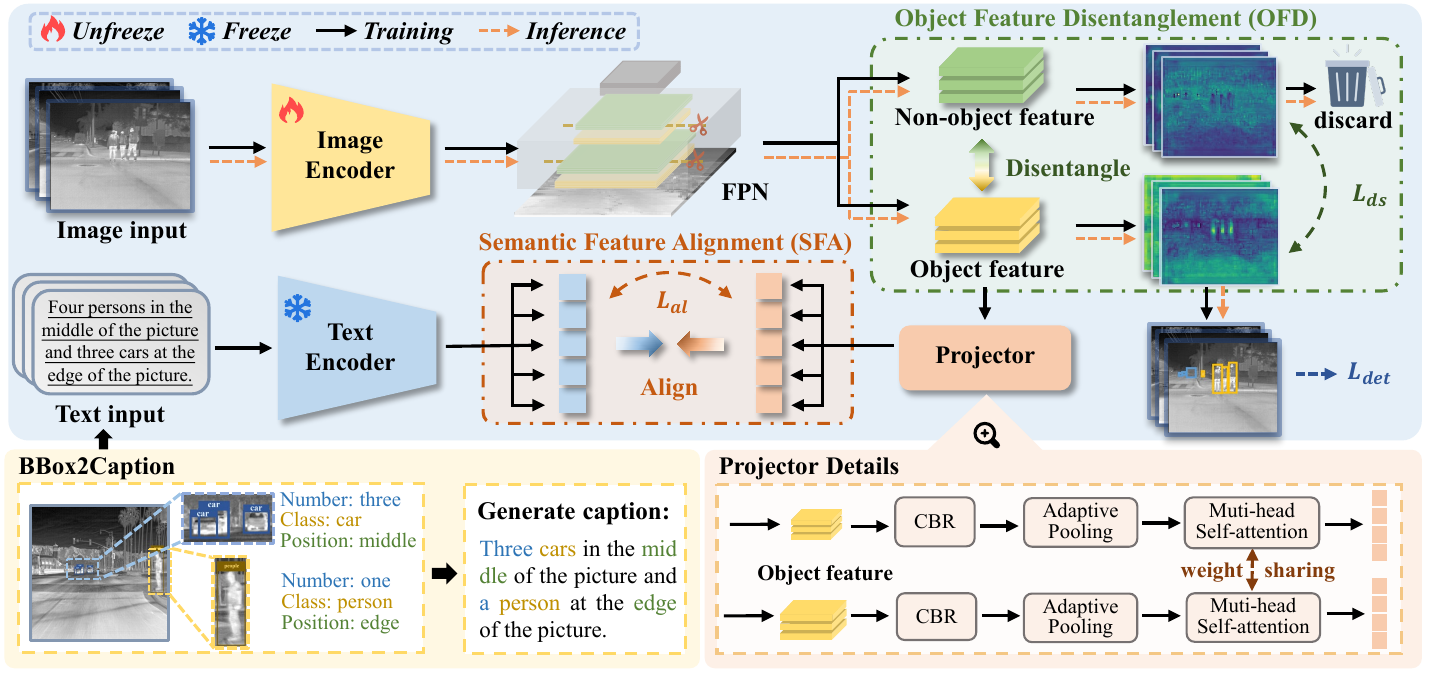}
    \caption{
    Our LGFD’s overall architecture. 
    Initially, BBox2Caption is introduced to generate the detailed descriptions in a rule-based manner. 
    The paired image-text data are entered into encoders to obtain features. Image FPN features are separated into object and non-object parts.
    Subsequently, the SFA module aligns object and text features to make the model learn semantic information pertinent to the objects.  
    Meanwhile, the OFD module minimizes the similarity between the object and non-object features via a constraint loss. It can mitigate the influence of background on object feature extraction.
    Ultimately, the robust object features are leveraged to produce final detection output.}
    \label{overview}
    \vspace{-0.3cm}
\end{figure*}
\section{Related Work}
\label{sec:relatedwor}
\subsection{Infrared Object Detection}
In recent years, object detection~\cite{yao2024domain} models have effectively balanced performance in terms of speed and accuracy, \eg, YOLO series~\cite{wang2024yolov10} and DETR~\cite{pu2024rank}. Some large-scale detectors~\cite{liu2023grounding,ren2024dino} have also  
demonstrated remarkable zero-shot detection performance. However, these methods~\cite{liu2025boost} were typically proposed for object detection in visible images.
Compared with well-developed visible 
domain, infrared object detection draws less attention from researchers.
A common paradigm is to fine-tune visible image pre-trained detectors on infrared data~\cite{RGBT,featurefusion-IRfinetune3}. Nevertheless, due to the domain gap and different image-forming principles between visible and infrared images, models tend to show inferior performance on infrared data.

To enhance infrared object detection~\cite{CrossFormer-RGB-IR-fusion2} and ~\cite{DEYOLO-RGB-IR-fusion1} developed multimodal networks that feed paired RGB and infrared samples into the network to detect objects in thermal images. However, obtaining paired images from both domains is difficult due to heavy manual image matching.
Therefore, there exited several methods that aim to perform IROD without additional paired RGB data~\cite{sharma2023tensor}.
For example, \cite{SMG-Y} proposed a transfer learning approach called Source Model Guidance (SMG), which utilized high-capacity RGB detection models to guide and supervise the training of infrared detection networks. \cite{YOLO-CIR-IRfinetune1} designed the coordinate attention module to focus on objects and suppress the background. Although the above efforts made several progress, IROD performance still can be further improved.

\subsection{Vision-language Representation Learning}
Vision-language models (VLMs) have demonstrated significant potential in integrating visual and linguistic information~\cite{VLM1,VLM2,VLM3,yao2025remotereasoner,yao2025remotesam}.
They applied cross-modal pre-training and contrastive learning to obtain discriminative representation. For example, CLIP~\cite{CLIP}, ALIGN~\cite{ALIGN}, and CyCLIP~\cite{goel2022cyclip} achieved notable success in zero-shot image recognition tasks. 
Regarding other visual and linguistic tasks,
Flamingo~\cite{alayrac2022flamingo} utilized the visual and language inputs as prompts and showed remarkable few-shot performance for visual question answering.
FLAVA~\cite{singh2022flava} advanced multi-modality understanding by integrating visual and linguistic information, thereby improving performance in image classification and text retrieval. In the area of multi-modal retrieval, ALIGN~\cite{ALIGN} illustrated how contrastive learning can establish strong associations between images and texts, providing new insights for image-text retrieval tasks. BLIP~\cite{BLIP} excelled in the interaction between images and texts, effectively conducting image-text matching and generation tasks.
Motivated by the success of the above works in the vision-language field, our approach applies a vision-language representation learning paradigm to enhance IROD performance.

\section{Proposed Method}

In this section, we elaborate the proposed Language-guided Feature Disentanglement (LGFD) method, which is a novel vision-language representation learning paradigm for IROD.
The overall framework is represented in Figure~\ref{overview}.
To illustrate the process of disentangling the object feature and the non-object feature, we provided the Algorithm ~\ref{alg1} of language guided feature disentanglement.

\subsection{Problem Formulation}
\subsubsection{Revisiting Infrared Object Detection}
The objective of the IROD task is to learn a mapping that generates the detection results $y_{i}$ from a set of input infrared samples $x_{i}$.This process can be formally represented as:
\begin{equation}
p(Y|X; \mathcal{M}_{\theta}) = \prod_{i=1}^{N} p(y_i | x_i ; \mathcal{M}_{\theta}),
\label{eq1}
\end{equation}
where $p(\cdot)$ represents the conditional probability. $X$ and $Y$ denote the input and output within the mapping, respectively. $N$ is the sample number.
$\mathcal{M}_{\theta}$ denotes the model with learnable parameters $\theta$. 

IROD typically involves classification and regression. The classification task identifies the category $c_i^k$ of each object: 
\begin{equation}
p(c_i^k \mid x_i; \mathcal{M}_\theta) = \Phi \left( \mathcal{F}(x_i; \mathcal{M}_\theta) \right)^k,
\label{eq2}
\end{equation}
where $k$ represents the $k^{th}$ object. $\Phi(\cdot)$ and $\mathcal{F}(\cdot)$ are the functions of classification and feature extraction, respectively. 

The regression task aims to accurately predict the location of each object.
Assuming that bounding box $b_i^k$ follows Gaussian distribution in the regression model~\cite{retinanet}, the process can be modeled as:
\begin{equation}
p(b_i^k \mid x_i; \mathcal{M}_\theta) = \mathcal{N} \left( b_i^k \mid \hat{b}_i^k (x_i; \mathcal{M}_\theta), \left( \sigma_i^k \right)^2 \right),
\label{eq3}
\end{equation}
where $k$ represents the $k^{th}$ object. $\hat{b}_i^k$ and $\sigma^2$ denote the predicted bounding box and variance.
The output $\hat y_i$ consists of both classification and regression predictions to provide the detection results:
\begin{equation}
\hat y_i = \left\{ p(c_i^k \mid x_i; \mathcal{M}_\theta) , p(b_i^k \mid x_i; \mathcal{M}_\theta)  \right\}.
\label{cls+reg}
\end{equation}

Then, the final optimization for IROD task is expressed as the following process:
\begin{equation}
\arg \min_{\theta} \mathcal{L}_{det} = {\sum_{i=1}^{N} {(d_{det}(\hat{y_i}, y_i)} ; \mathcal{M}_\theta)},
\label{eq4}
\end{equation}
where $d_{det}(\cdot)$ is defined as distance metric functions between predictions and ground truth labels.

\subsubsection{Language-guided Detection Paradigm}
Our proposed LGFD separates the object and non-object features through language guidance. 
By integrating text supervision, the typical infrared detection pipeline is transformed into an infrared vision-language representation learning paradigm. 
The input is expanded from the original single image $x_i$ to a binary $(x_{i},t_{i})_{i=1}^{N}$, where $t_{i}$ is the auxiliary caption. Then, the mapping process of LGFD can be expressed as follows:
\begin{equation}
p(Y|(X,T); \mathcal{M}_{\theta}) = \prod_{i=1}^{N} p(y_i | (x_i,t_i) ; \mathcal{M}_{\theta}),
\label{eq5}
\end{equation}
where $X$, $T$, and $Y$ represent the visual input, textual input, and detection results within the mapping, respectively.

LGFD disentangles the visual features into object and non-object features prior to detection head as:
\begin{equation}
f_i ^{obj}, f_i ^{nobj} = g(\mathcal{F}(x_i, t_i; \mathcal{M}_\theta)),
\label{eq6}
\end{equation}
where $f_i ^{obj}$ and $ f_i ^{nobj}$ are object and non-object features. $g(\cdot)$ denotes the disentangle function. The decoupled object features $f_i ^{obj}$ are entered into the detection head for subsequent predictions $\hat{y_i}$. Consequently, the optimization objective with parameters $\theta$ transitions from~\eqref{eq4} to~\eqref{eq7}:

\begin{equation}
\small
\arg\min_{\theta} \mathcal{L} = {\sum_{i=1}^{N} ( (d_{det}(\hat{y_i}, y_i) + d_{dis}(f_i ^{obj}, f_i ^{nobj}) ; \mathcal{M}_\theta)},
\label{eq7}
\end{equation}
where $d_{dis}(\cdot)$ is defined as distance metric functions between the extracted features, such as the cosine similarity or Euclidean distance. Minimizing the  $\mathcal{L}$ effectively integrates IROD performance enhancement and object feature disentanglement.

\subsection{Auxiliary Captions Generation}
\label{sec3.2}
Previous IROD methods are merely supervised by detection labels (categories and bounding boxes). Differently, our approach explores the possibility of leveraging supervisory information in an alternative manner.Specifically, we utilize infrared image annotations to automatically generate descriptive captions, as depicted in Figure~\ref{overview}.

We adopt a rule-based approach, namely Bbox2Caption, to transfer the bounding box annotations into a set of natural language captions.
Specifically, it initially computes the number of objects and subsequently integrates number, category, and spatial information to generate a caption that provides detailed object descriptions. 
When an object appears more than 10 times, a more generalized term (\eg, `a large number of', `lots of') is employed in place of the exact number to enhance the readability and variability of the captions. Finally, we generate an overall caption for each image, which contains detailed information of objects.

\begin{algorithm}[t]
\caption{\textbf{
LGFD
}}
\raggedright  
\textbf{Require:} \\  
captions: Text data that matches the images,\\
$\mathcal{M}_{\theta}$: Detector with parameter $\theta$, \\
$f_{i}^{ori} \in \mathbb{R}^{H \times W \times 2L}$: Feature map of $P_{3}$ and $P_{4}$ ,\\
$\mathcal{L}_{det},\mathcal{L}_{al},\mathcal{L}_{ds}$: The object detection loss, the alignment loss and the disentanglement loss. \\
    \begin{algorithmic}
        \FOR{each training epoch}
            \FOR{each training sample feature 
            $x_{\mathcal{M}_{\theta}}^{ori}$}
                \STATE  \textit{\textcolor{blue}{\# Decomposing feature maps of $P_{3}$ and $P_{4}$.}} \\  $f^{obj}_i, f^{nobj}_i = \text{Decompose}(f^{ori}_i;\mathcal{M}_\theta)$ \\
                \STATE
                \textit{\textcolor{blue}{\# Obtain the textual feature.}}
                \\ $f_i^{text}= BERT(captions)$  
                \STATE 
                 \textit{\textcolor{blue}{\# Projecting $f^{obj}_i$ for calculating the similarity.}}
                 \\$\hat f_i^{obj}= Projector(f_i^{obj})$\\
                  $S_{ij} = \frac{\hat f_i^{obj} \cdot f^{text}_j}{\| \hat f_i^{obj} \| \| f^{text}_j \|}$\\
                 \STATE 
                 \textit{\textcolor{blue}{\# Optimize $\mathcal{L}$ to find optimal $f_i^{obj}$.}}\\
                 $\mathcal{L}_{ds} = cos(Pooling(f^{obj}) ,Pooling(f^{nobj}))$ \\
                 $\mathcal{L}_{al} = - \frac{1}{b} \sum_{i=1}^{b} \log \frac{\exp(S_{ii}/\tau)}{\sum_{j=1}^{b} \exp(S_{ij}/\tau)}$ \\
                 $\mathcal{L} = \mathcal{L}_{det} + \alpha \mathcal{L}_{al} + \beta \mathcal{L}_{ds}$ \\
            \ENDFOR
        \ENDFOR
        \RETURN Optimized detecotr $\mathcal{M}_{\theta}$
    \end{algorithmic}
\label{alg1}
\end{algorithm}

\subsection{Infrared Object Feature Disentanglement}
The object and non-object features entangled in the feature space can decrease the model performance. Therefore, we aim to disentangle them with the guidance of textual semantic information to facilitate detection.

\subsubsection{Visual Feature Decomposition}
We perform channel decomposition directly on the Feature Pyramid Network (FPN), partitioning the visual features extracted by the backbone into two distinct groups. The decomposition process can be expressed as:
\begin{equation}
f^{obj}_i, f^{nobj}_i = \text{Decompose}(f^{ori}_i; \mathcal{M}_\theta),
\label{decompose}
\end{equation}
where $f^{obj}_i \in \mathbb{R}^{H \times W \times L} $ and $f^{nobj}_i \in \mathbb{R}^{H \times W \times L}$ are the two feature maps decomposed from $f^{ori}_i \in \mathbb{R}^{H \times W \times 2L}$. $f^{ori}_i$ is the initial feature extracted by the model $\mathcal{M}_\theta$. 
Subsequently, we should disentangle $ f^{obj}_i$ and $f^{nobj}_i$ as completely as possible, making them learn object and non-object features respectively.
Afterward, the object features will be utilized for robust infrared detection tasks, while the non-object features should be discarded as task-independent components.

\subsubsection{Semantic Feature Alignment}
We employ the contrastive learning strategy to make $ f^{obj}_{i}$ learn object feature in an interpretable manner. 
Specifically, the visual backbone is trained with image-text dataset $\mathcal{D} = {(x_{i}, t_{i})}_{i=1}^{N}$ where the auxiliary text $t_{i}$ is automatically generated as described in Section~\ref{sec3.2}. Then, we apply a pre-trained text encoder (\eg, BERT~\cite{bert}) to obtain text features $f_i^{text} \in \mathbb{R}^{1 \times L}$.

Contrastive learning~\cite{CLIP} requires the dimensionality alignment between visual and textual features. 
Therefore, we design a projector to map the dimensionalities of decomposed object features $ f^{\text{obj}}$ and text features $f_i^{text}$:
\begin{equation}
\hat f_i^{obj} = MutiHead(Pooling(CBR(f^{\text{obj}}))),
\label{projector}
\end{equation}
where $CBR$ represents a block composed of $1 \times  1$ convolution layer ($C$), a batch normalization layer ($B$) and a rectified linear unit ($R$). Then, an adaptive pooling is employed to dynamically adjust the feature dimensions. Next, a multi-head self-attention layer is adopted to capture complex dependencies. Ultimately, we can obtain projected object features $\hat f_i^{obj}$ with dimensionality aligned to text features $f_i^{text}$. 

Subsequently, the contrastive loss is utilized to align the dimension-matched object and text features in a unified feature space. This loss function facilitates $\hat f_i^{obj}$ to learn the object relevant features from textual semantic information.
Specifically, $f_i^{text}$ and $\hat f_i^{obj}$ are concatenated within a training batch of size $b$, forming two embedding groups $\{ \hat f_1^{obj}, \hat f_2^{obj}, \dots, \hat f_b^{obj} \} \{ f_1^{text}, f_2^{text}, \dots, f_b^{text} \} \in \mathbb{R}^{b \times L}$. Afterwards, we compute cosine similarities between visual and textual embedding groups:

\begin{equation}
S_{ij} = \frac{\hat f_i^{obj} \cdot f^{text}_j}{\| \hat f_i^{obj} \| \| f^{text}_j \|},  
\label{similarity1}
\end{equation}
where $S_{ij}$ denotes the similarity of $i^{th}$ visual feature and $j^{th}$ textual feature.
Then, the contrastive loss is calculated on the similarity matrix, which prompts pull the matched image-text samples while pushing apart the mismatched samples. The loss is given by the following equation:
\begin{equation}
\mathcal{L}_{al} = - \frac{1}{b} \sum_{i=1}^{b} \log \frac{\exp(S_{ii}/\tau)}{\sum_{j=1}^{b} \exp(S_{ij}/\tau)},
\label{infonce}
\end{equation}
where $\tau$ is the temperature parameter. Ultimately, $f_i^{obj}$ can effectively learn object features by establishing a strong correlation between visual features and the semantic information from auxiliary captions.

\subsubsection{Object Feature Disentanglement}
Although our SFA module ensures $f^{obj}_i$ to learn the object features, $f^{nobj}_i$ may still reserve a portion of the object features.
In this situation, if $x^{nobj}$ is discarded, it may negatively impact the detection performance.
In order to guide $f^{obj}_i$ to learn intact object features, an intuitive idea is to push $f^{obj}$ and $f^{nobj}$ apart to disentangle them.
To accomplish this objective, we employ a constraint on $f^{obj}$ and $f^{nobj}$ to minimizes their similarity:
\begin{equation}
\mathcal{L}_{ds} = cos(Pooling(f^{obj}) ,Pooling(f^{nobj})),
\label{disentangle_loss}
\end{equation}
where $cos$ means the cosine similarity. An average pooling is adopted to reduce the dimensionality.  
In this manner, object and non-object features can be significantly disentangled in the feature space. 

Finally, considering the alignment loss $\mathcal{L}_{al}$, the disentanglement loss $\mathcal{L}_{ds}$, and the object detection loss $\mathcal{L}_{det}$.
Our overall training objective can be expressed as:
\begin{equation}
\mathcal{L} = \mathcal{L}_{det} + \alpha \mathcal{L}_{al} + \beta \mathcal{L}_{ds},
\label{loss_total}
\end{equation}
where $\alpha$ and $\beta$ represent the hyper-parameters to balance each loss. 

The captions are auxiliary supervision in the training phase to enhance object feature extraction. 
In the inference, no additional caption information is required.
Throughout both phases, the disentangled object features are fed into the detection head to produce final results, while non-object features are simply discarded. Furthermore, since non-object features are discarded, the model inference complexity slightly decreases. 
\begin{table}[t]
\caption{Comparison on FLIR dataset with infrared-visible and infrared-only approaches.}
\centering
\renewcommand\arraystretch{1.05}
\setlength{\tabcolsep}{2.2mm}{
\begin{tabular}{lc|cc}
\toprule
Method                   & Backbone  & $mAP$    & $AP_{50}$\\
\midrule          
\cellcolor[HTML]{E0E0E0}\textit{Infrared-visible} &\cellcolor[HTML]{E0E0E0}&\cellcolor[HTML]{E0E0E0}&\cellcolor[HTML]{E0E0E0}\\
GAFF (2021)        &ResNet18        & 37.5        & 72.9\\          
GAFF (2021)          & VGG16            &37.3       & 72.7\\   
CFT (2021)           & CFB             & 40.2        & 78.7\\   
IGT (2023)         & Swin-T-Tiny        & 43.6       & 85.0\\
CrossFormer (2024)   & ViT             & 42.1        & 79.3\\         
YOLO-Fusion (2024)   & CSPDarknet       & -        & 84.7\\           
FD\textsuperscript{2}-Net (2024)   & CSPDarknet       & -      & 82.9\\
\midrule      
\cellcolor[HTML]{E0E0E0}\textit{Infrared-only}   &\cellcolor[HTML]{E0E0E0}  &\cellcolor[HTML]{E0E0E0}&\cellcolor[HTML]{E0E0E0}\\
SSD (2016)        & VGG16       & 29.6       & 65.5\\
Faster-RCNN (2016)      & ResNet50      & 37.6       & 74.4\\
SMG-Y (2022)   & -      & -       & 77.0\\
YOLOv5m (2020)       & CSPDarknet       & -       & 81.9\\
YOLO-ACN (2020)       & CSPDarknet      & -       & 82.3\\
RGBT (2022)      & -       & -       &82.9\\
YOLO-CIR (2023)       & CSPDarknet      & -       & 84.9\\
\cellcolor[HTML]{DAE8FC}LGFD (ours)        &  \cellcolor[HTML]{DAE8FC}CSPDarknet      & \cellcolor[HTML]{DAE8FC}\textbf{45.8 }       & \cellcolor[HTML]{DAE8FC}\textbf{86.1}\\
\bottomrule
\end{tabular}
}
\vspace{-0.4cm}
\label{tab:comparison-FLIR}
\end{table}
\section{Experiments}

\subsection{Experimental Settings}
\textbf{Datasets.} The proposed model is evaluated on two publicly available datasets:
(1) \textbf{FLIR} offers a challenging infrared object detection benchmark. We utilized the aligned version with more precise labeling as provided by~\cite{flir-aligned}. The dataset includes 5,142 precisely aligned infrared-visible paired images with 4,129 and 1,013 samples assigned for training and testing, respectively. 
(2) \textbf{M\textsuperscript{3}FD} comprises 4,200 pairs of visible and thermal images captured by on-board cameras~\cite{TarDAL}. 
Following the original setup of the dataset, we utilized 80\% of the images for training and the remaining 20\% images for evaluation.

Importantly, since we concentrated on investigating IROD without paired RGB images, we only selected the infrared data in the above datasets for training.

\noindent\textbf{Experimental Details.}
We selected YOLOv7-L~\cite{yolov7} as the base detector for our proposed LGFD. It is initialized with the MS-COCO\cite{mscoco} pre-trained weight. We utilized a widely-used BERT~\cite{bert} model as our text encoder.
To evaluate object detection performance, we selected the widely-adopted metrics, mean average precision $mAP$ and average precision at 50 IoU threshold $AP_{50}$. 
Our framework is trained for 300 epochs with a batch size of 16. 
We fixed the random seed to 42 to ensure the reproducibility of the experiments. All experiments are conducted on an NVIDIA RTX 3090 GPU.

\subsection{Main Results}

\begin{figure*}[t]
    \centering
    \includegraphics[width=0.99\linewidth]{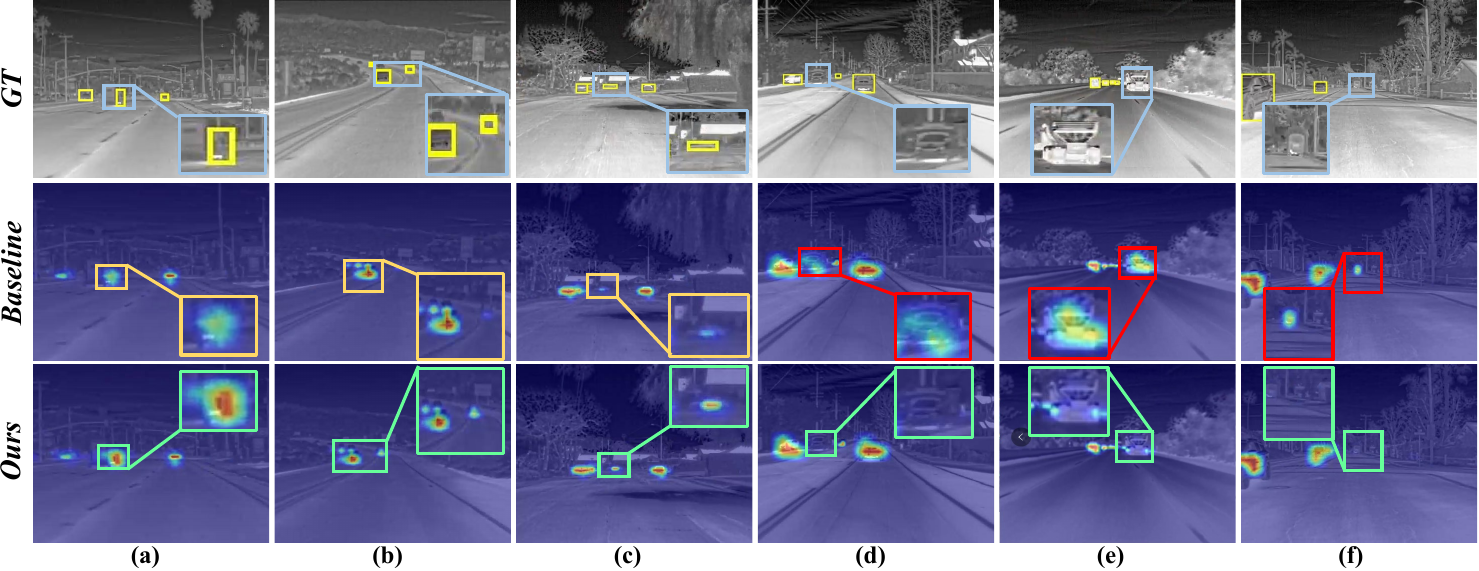}
    \caption{Visualization of feature activations results of our LGFD and baseline on FLIR with ground truth (GT). Orange (a-c) and red (d-f) boxes indicate weak and false activations, respectively. Green boxes represent the results of LGFD for comparison.}
    \label{fig3}
\end{figure*}

\textbf{Compared Methods.}
We compare our proposed LGFD with several SOTA works, including infrared-visible and infrared-only methods.
\textbf{Infrared-visible} object detection approaches utilize paired visible and infrared images, including GAFF~\cite{gaff}, CFT~\cite{CFT}, CrossFormer~\cite{CrossFormer-RGB-IR-fusion2}, YOLO-Fusion~\cite{YOLO-FUSION}, IGT~\cite{IGT}, FD\textsuperscript{2}-Net~\cite{fd2net}, RFN~\cite{RFN}, DeFusion~\cite{defusion}, TarDAL~\cite{TarDAL}, IGNet~\cite{IGNET}, MetaF~\cite{metafusion}. 
\textbf{Infrared-only} object detection methods merely leverage infrared samples without paired RGB data, including specifically SSD~\cite{ssd}, Faster-RCNN~\cite{fasterrcnn}, SMG-Y~\cite{SMG-Y}, YOLO-ACN~\cite{yolo-acn}, RGBT~\cite{RGBT}, YOLO-CIR~\cite{YOLO-CIR-IRfinetune1}, CenterNet2~\cite{centernet2}, Sparse RCNN~\cite{sparsercnn}, YOLOv7-tiny~\cite{yolov7}, Swin Transformer~\cite{swintransformer}.



\begin{table}
\caption{Comparison on M\textsuperscript{3}FD dataset with infrared-visible and infrared-only methods.}
\centering
\renewcommand\arraystretch{1.05}
\setlength{\tabcolsep}{1.5mm}{
\begin{tabular}{lc|cc}
\toprule
Method      & Backbone     & $mAP$  & $AP_{50}$\\
\hline
\cellcolor[HTML]{E0E0E0}\textit{Infrared-visible} &\cellcolor[HTML]{E0E0E0}   &\cellcolor[HTML]{E0E0E0}   &\cellcolor[HTML]{E0E0E0} \\
RFN (2022)          &  CSPDarknet        &53.2     & 79.4 \\
TarDAL (2022)       &  CSPDarknet        &54.1     & 80.6  \\
DeFusion (2022)     &  CSPDarknet          &54.1      & 80.8 \\
IGNet (2023)        &  CSPDarknet       & 54.5     & 81.5 \\
MetaF (2023)        &   CSPDarknet       & 56.5       & 81.6   \\
FD\textsuperscript{2}-Net (2024)        &   CSPDarknet       & -       & 83.5   \\
\hline
\cellcolor[HTML]{E0E0E0}\textit{Infrared-only} &\cellcolor[HTML]{E0E0E0}       &\cellcolor[HTML]{E0E0E0}   &\cellcolor[HTML]{E0E0E0} \\
Swin Transformer (2021)           & -       & 41.9   & 72.6\\
CenterNet2 (2021)           & ResNet50       & 42.4   & 65.3\\
Sparse RCNN (2021)          & ResNet50            & 44.8   & 76.4   \\
YOLOv7-tiny (2023)          & CSPDarknet       & 48.4   & 78.1\\

\cellcolor[HTML]{DAE8FC}LGFD (ours)        &\cellcolor[HTML]{DAE8FC}CSPDarknet       &\cellcolor[HTML]{DAE8FC} \textbf{51.7}   &\cellcolor[HTML]{DAE8FC}\textbf{83.7} \\
\bottomrule
\end{tabular}
}
\vspace{-0.4cm}
\label{tab:comparison on m3fd}
\end{table}

\noindent\textbf{Experiments on FLIR.}
We validated the effectiveness of LGFD on the FLIR dataset. From the experimental results in Table~\ref{tab:comparison-FLIR}, it can be observed that in comparison to infrared-only methods, ours obtains a 0.8\% improvement in $AP_{50}$ over the suboptimal method. 
More importantly, our method surpasses all infrared-visible methods. 
Specifically, compared to IGT~\cite{IGT}, LGFD achieves significant performance gains of 2.7\% and 0.7\% on $mAP$ and $AP_{50}$, respectively. Considering that our approach merely utilizes infrared images without paired visible data, its superior performance over infrared-visible methods significantly demonstrates the advantages of our approach.

To directly explain the superior performance of our approach, we visualized the feature activation map compared with the baseline detector (YOLOv7-L) in Figure~\ref{fig3}.
The results in columns (a-c) represent that our method exhibits stronger activations to the interest regions where the ground truth exists.
Moreover, the baseline has false activations in Figure~\ref{fig3} (d-f), while ours overcomes this drawback. The reason is that our LGFD can discard non-object features via disentanglement, thereby reducing the risk of being misguided by false feature activations.
To sum up, stronger true and fewer false feature activations can account for the significant performance improvements of our approach.


\noindent\textbf{Experiments on M\textsuperscript{3}FD.} We presented the experimental results in Table~\ref{tab:comparison on m3fd}.
It can be observed from Table~\ref{tab:comparison on m3fd} that our proposed LGFD surpasses all mainstream state-of-the-art (SOTA) methods in $AP_{50}$.
Specifically, compared with infrared-only methods, LGFD shows advantages over the second-best method in $AP_{50}$ by 5.4\% and in $mAP$ by 2.7\%.
In comparison to infrared-visible detection frameworks, our LGFD also exhibits superior performance in the $AP_{50}$. Although it is inferior to some infrared-visible methods in $mAP$, we believe this phenomenon is reasonable because LGFD doesn't utilize visible images as auxiliary data. 

The experiments on two datasets indicate that the proposed approach can effectively enhance infrared object detection performance. Although it merely utilizes infrared data, it could even surpass infrared-visible detection methods. This satisfying result reveals the effectiveness of our vision-language representation learning paradigm and feature disentangling strategy for IROD.

\begin{table}[t]
\caption{Ablation study of LGFD components. SFA: Semantic feature alignment. OFD: Object feature disentanglement.}
\centering
\renewcommand\arraystretch{0.975}
\setlength{\tabcolsep}{0.9mm}{
\begin{tabular}{c|cc|ccc}
\toprule
Dataset & SFA  & OFD   & $mAP$     & $AP_{50}$  & $AP_{75}$\\
\hline
\multirow{4}{*}{FLIR}
&                  &               & 44.2  &  84.2    & 39.5 \\
&   \Checkmark     &               & 45.3 (+1.1)  &  85.7 (+1.5)    & 40.6 (+1.1) \\
&                  &    \Checkmark & 45.2 (+1.0)  &  85.4 (+1.2)    & 39.7 (+0.2) \\

&  \cellcolor[HTML]{DAE8FC}\Checkmark     & \cellcolor[HTML]{DAE8FC}\Checkmark   &\cellcolor[HTML]{DAE8FC}\textbf{45.8 (+1.6)}  & \cellcolor[HTML]{DAE8FC}\textbf{86.1 (+1.9)}    & \cellcolor[HTML]{DAE8FC}\textbf{41.1 (+1.6)} \\
\hline
\multirow{4}{*}{M\textsuperscript{3}FD}
&                  &               & 47.6  &  78.3    & 49.0 \\
&   \Checkmark     &               & 50.2 (+2.6)  &  82.9 (+4.6)    & 52.8 (+3.8) \\
&                  &    \Checkmark & 51.1 (+3.5)  &  83.3 (+5.0)    & 53.0 (+4.0) \\
& \cellcolor[HTML]{DAE8FC}\Checkmark     & \cellcolor[HTML]{DAE8FC}\Checkmark   &\cellcolor[HTML]{DAE8FC}\textbf{51.7 (+4.1)}  &\cellcolor[HTML]{DAE8FC}\textbf{83.7 (+5.4)}    &\cellcolor[HTML]{DAE8FC}\textbf{53.3 (+4.3)} \\
\bottomrule
\end{tabular}
}
\vspace{-0.4cm}
\label{tab:ablation-afs-ofd}
\end{table}

\begin{figure*}[t]
    \centering
    \includegraphics[width=0.99\linewidth]{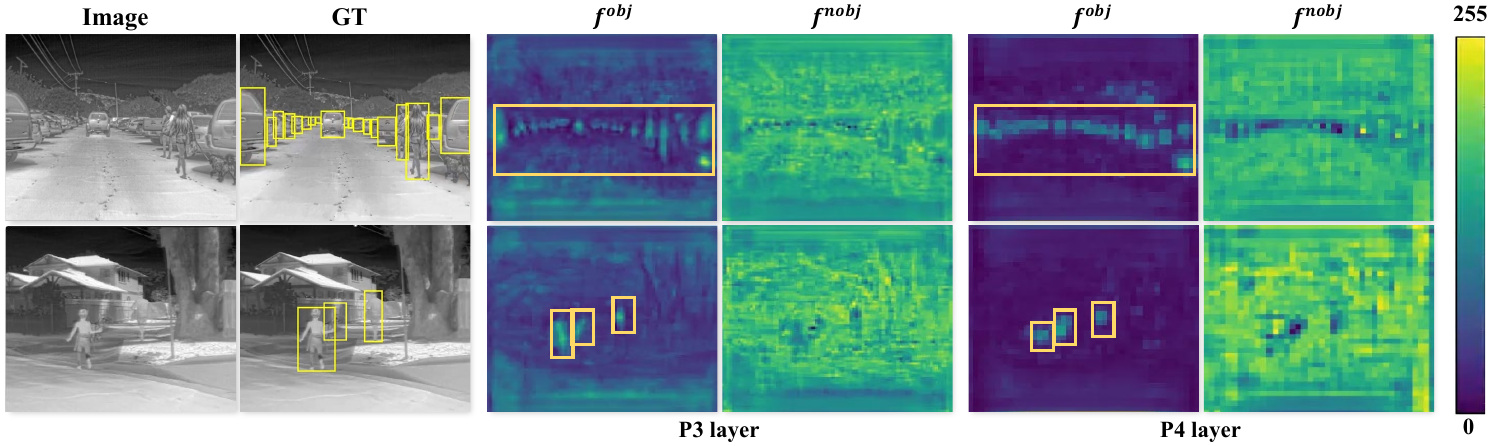}
    \caption{Visualization of object and non-object features of P3 and P4 layers in FPN. The object features are more focused on the foreground where the ground truth (GT) exists. Conversely, non-object features are primarily concerned with the background.}
    \label{fig4}
\end{figure*}

\subsection{Ablation Studies}

\subsubsection{Effectiveness of SFA and OFD Modules}
Our method LGFD consists of two key components: Semantic Feature Alignment (SFA) and Object Feature Decoupling (OFD). Table~\ref{tab:ablation-afs-ofd} presents the effects of them for the final detection performance. It can be observed that the $AP_{50}$ increases by 1.9\% and 5.4\% on the FLIR and M\textsuperscript{3}FD respectively, when employing the SFA module. The incorporation of the OFD module further enhanced performance, leading to additional gains of 0.4\% and 0.8\% in $AP_{50}$ on the two datasets. These results indicate that learning object features with language guidance significantly contributes to the final improvement (SFA).
Additionally, the subsequent integration of disentangling object and non-object features can further boost the performance (OFD).

\begin{table}[t]
\caption{Impact of utilizing object features $f^{obj}$ and non-object features $f^{nobj}$ with YOLOv7 as the base detector.}
\centering
\renewcommand\arraystretch{0.95}
\setlength{\tabcolsep}{0.9mm}{
\begin{tabular}{c|cc|ccc}
\toprule
Dataset & $f^{obj}$  & $f^{nobj}$   & $mAP$     & $AP_{50}$  & $AP_{75}$\\

\hline
\multirow{4}{*}{FLIR}
&                  &               & 44.2  &  84.2    & 39.5 \\
&  \cellcolor[HTML]{DAE8FC}  \Checkmark     &  \cellcolor[HTML]{DAE8FC}              & \cellcolor[HTML]{DAE8FC}\textbf{45.8 (+1.6)}  & \cellcolor[HTML]{DAE8FC}\textbf{86.1 (+1.9)}    & \cellcolor[HTML]{DAE8FC}\textbf{41.1 (+1.6)} \\
&                  &    \Checkmark & 44.3 (+0.1)  &  84.6 (+0.4)    & 39.2 (-0.3) \\
&   \Checkmark     &  \Checkmark   & 44.4 (+0.2)  &  82.8 (-1.4)    & 39.6 (+0.1) \\

\hline
\multirow{4}{*}{M\textsuperscript{3}FD}
&                  &               & 47.6  &  78.3    & 49.0 \\
&  \cellcolor[HTML]{DAE8FC}\Checkmark     &  \cellcolor[HTML]{DAE8FC}              & \cellcolor[HTML]{DAE8FC}\textbf{51.7 (+4.1)}  & \cellcolor[HTML]{DAE8FC}\textbf{83.7 (+5.4)}    & \cellcolor[HTML]{DAE8FC}\textbf{53.3 (+4.3)} \\
&                  &    \Checkmark & 49.3 (+1.7)  &  81.7 (+3.4)    & 51.1 (+2.1) \\

&   \Checkmark     &  \Checkmark   & 51.3 (+3.7)  &  81.8 (+3.5)    & 52.1 (+3.1) \\
\bottomrule
\end{tabular}
}
\vspace{-0.4cm}
\label{tab:ablation-nobj-features}
\end{table}

\subsubsection{Impact of Object and Non-object Features}
To obtain deeper insights into the disentangled features, 
we conducted experiments to validate the impact of object and non-object features on detection.Specifically, 
we enter the object feature and both object and non-object features into the detection head, respectively. As demonstrated in Table~\ref{tab:ablation-nobj-features}, utilizing $f^{obj}$ effectively enhanced the performance by 5.4\% and 1.9\% of $AP_{50}$ on two datasets compared to the baseline. However, if utilizing both $f^{obj}$ and $f^{nobj}$ features, the improvement over the baseline was decreased. The $AP_{50}$ reductions are 3.3\% and 1.9\% on the FLIR and M\textsuperscript{3}FD respectively.
This result indicates that the non-object feature may contain detrimental information and discarding it can enhance the performance.

\subsubsection{Feature Disentanglement Visualization}
To investigate the feature representation capabilities of the proposed LGFD,we conducted a visual analysis of the object and non-target features after disentangling the P3 and P4 layers of FPN. 
As illustrated in Figure~\ref{fig4}, it is evident that LGFD significantly disentangles the features into object and non-object components. 
The high-response regions of $f^{obj}$ and $f^{nobj}$ correspond to the foreground and background respectively.
Therefore, the effectiveness of LGFD lies in its ability to concentrate on robust object features while suppressing background noise.

\subsubsection{Decomposement at Different FPN Layers}
We explore the effectiveness of disentanglement at different locations through experiments conducted on various resolution layers of the FPN. 
The results are presented in Table~\ref{tab:ablation-FPN}, and the best performance are achieved by disentangling at the P3 and P4 layers.
The reason could be that
the proportion of large-scale objects, which P5 layers detects, in the two datasets is relatively small, accounting for about 7\% and 11\%, respectively.
Consequently, we ultimately choose to disentangle the P3 and P4 layers that correspond to detect small and medium-scale objects, with a significant improvement of 1.9\% on the P3 and P4 layers.

\begin{table}[t]
\caption{The impact of decomposing different layers in FPN for the detection performance on FLIR dataset.}
\centering
\renewcommand\arraystretch{1.2}
\setlength{\tabcolsep}{1.1mm}
\begin{tabular}{c|c|cccccccc}
\toprule
\multirow{3}{*}{FPN}
& P3    &       & \Checkmark    &       &     &\cellcolor[HTML]{DAE8FC} \Checkmark    & \Checkmark    &      & \Checkmark\\
& P4    &       &           & \Checkmark     &     &\cellcolor[HTML]{DAE8FC} \Checkmark        &   & \Checkmark    & \Checkmark\\
& P5    &       &     &      & \Checkmark    &\cellcolor[HTML]{DAE8FC}         & \Checkmark  & \Checkmark    & \Checkmark\\
\hline
\multicolumn{2}{c|}{$AP_{50}$}  & 84.2      & 85.5      & 85.4      & 83.8      &\cellcolor[HTML]{DAE8FC} \textbf{86.1}      & 85.2      & 84.2      & 84.0  \\
\hline
\multicolumn{2}{c|}{$\pm \Delta$}  & -       & +1.3      & +1.2      & -0.4      &\cellcolor[HTML]{DAE8FC} \textbf{+1.9}      & +1.0      & 0.0      & -0.2      \\
\bottomrule
\end{tabular}
\label{tab:ablation-FPN}
\end{table}


\subsubsection{Different Caption Generation Approaches}
To select a proper caption generation method, we compared the Large Vision-Language Model (LVLMs-based) by Qwen-VL~\cite{Qwen-VL} and the rule-based method.
The results are illustrated in Table~\ref{tab:ablation-qwencaption}. We can observe that text supervision is effective. Furthermore, the rules-based method surpasses LVLMs-based method by 0.6\% and 0.3\% on $AP_{50}$ on the FLIR and M\textsuperscript{3}FD, respectively.
This phenomenon is mainly attributed to the fact that the typical LVLM is not optimized for infrared modality and the hallucination problem tends to provide fake information.
Therefore, we adopt simple yet effective rule-based caption generation.

\begin{table}[t]
\caption{The impact of different caption generation strategies on the two datasets.}
\centering
\renewcommand\arraystretch{1.05}
\setlength{\tabcolsep}{2.9mm}{
\begin{tabular}{c|c|ccc}
\toprule
Dataset & Method   & $mAP$     & $AP_{50}$  & $AP_{75}$\\
\hline
\multirow{4}{*}{FLIR}
&  no-caption     & 44.2         &  84.2           & 39.5 \\
&  LVLMs-based    & 45.3         & 85.5            & 40.8 \\
&  \cellcolor[HTML]{DAE8FC}\textbf{rule-based}     & \cellcolor[HTML]{DAE8FC}\textbf{45.8}  &  \cellcolor[HTML]{DAE8FC}\textbf{86.1}    & \cellcolor[HTML]{DAE8FC}\textbf{41.1} \\
\hline
\multirow{4}{*}{M\textsuperscript{3}FD}
&   no-caption     & 47.6         &  78.3           & 49.0 \\
& LVLMs-based  &51.2  &83.4    &53.1 \\
&   \cellcolor[HTML]{DAE8FC}\textbf{rule-based}    & \cellcolor[HTML]{DAE8FC}\textbf{51.7}  &  \cellcolor[HTML]{DAE8FC}\textbf{83.7}    & \cellcolor[HTML]{DAE8FC}\textbf{53.3} \\
\bottomrule
\end{tabular}
}
\label{tab:ablation-qwencaption}
\end{table}
\section{Conclusion}
In this paper, we proposed a Language-Guided Feature Disentanglement (LGFD) approach specifically designed for infrared object detection tasks. It effectively separates object and non-object features by incorporating semantic language guidance. SFA module achieves effective feature disentanglement through the alignment of object features with corresponding textual features.
OFD module ensures that the model can focus on object features while minimizing the impact of non-object features. 
By converting the detection pipeline into a vision-language representation learning framework, LGFD significantly enhances infrared object detection performance. Experiments demonstrate the effectiveness of integrating semantic information from text modality. In future work, we will further investigate the potential of LGFD in other multimodal perception tasks.

{
    \small
    \bibliographystyle{ieeenat_fullname}
    \bibliography{main}
}

\setcounter{section}{0}
\clearpage
\setcounter{page}{1}
\maketitlesupplementary


\section{Additional Ablation Studies} \label{ablation}
\subsection{Effect of Balance Factors} 
\label{ablation-balance}
To ascertain the optimal balance factor values for model training,
we conducted experiments on M\textsuperscript{3}FD with different weights for the hyper-parameter $\alpha$ and $\beta$. 
The results are demonstrated in Table~\ref{tab:ablaion-balabce}. 
Models were trained using various parameter settings and the mean average precision ($mAP$) was recorded at the $50^{th}$ epoch to save experiment time.
It is evident that increasing or decreasing the weight $\alpha$ of aligning loss $\mathcal{L}_{al}$ significantly impacts the model's detection performance. This result reflects the importance of learning object features from the textual information.

The accuracy tends to decrease when the weight $\beta$ of disentangling loss $\mathcal{L}_{ds}$ exceeds $ \alpha $.
The reason may be that excessive disentanglement can weaken the strength of learning object features from semantic language guidance. 
Additionally, the detection accuracy may also be inferior when the disparity between the two balance factors is relatively large. 
This phenomenon suggests that balanced weights allocation can lead to superior model performance. Consequently, we ultimately adopt the set of balance factors of the highest accuracy in LGFD.

\subsection{Anlysis of CBR Block in Projector}
We conducted experiments on the FLIR and M\textsuperscript{3}FD datasets to validate the optimal choice of projector, including the number and size of convolution kernels in the CBR blocks. The results are demonstrated in Table~\ref{tab:ablation-cbr-flir} and Table~\ref{tab:ablation-cbr-m3fd}.
We selected CBR blocks with convolution kernel sizes of 1 and 3 to align feature dimensions (channels) for feature extraction. 
The results indicated that our LGFD achieved the best performance with a 1×1 convolution kernel in CBR block. Compared to scenarios without CBR blocks, this approach yielded performance gains of 2.3\% and 0.5\% in $AP_{50}$ on the FLIR and M\textsuperscript{3}FD datasets, respectively.
Therefore, in all experiments we utilized CBR blocks with a 1×1 convolution kernel to achieve optimal performance.

\subsection{Detailed Detection Results}
We compared the performance of our proposed method LGFD with the base detector YOLOv7-L as baseline on two IROD benchmark datasets. The detailed results are demonstrated in Table~\ref{tab:comparison-baseline} with $AP_{S}$, $AP_{M}$, and $AP_{L}$ as additional elevation metrics.

From the results in Table~\ref{tab:comparison-baseline}, we can observed that $AP_{S}$ to $AP_{L}$ all rises on two datasets. This phenomenon means that our approach can consistently improve the detection result across different scales. Specifically, on the M\textsuperscript{3}FD dataset, the detection performance on small objects ($AP_{S}$) remarkably enhanced by +6.4\%. Since small objects tend to be more difficult to detect, the improvement on them can support the effectiveness of our approach.

\setcounter{table}{0}
\begin{table}[t]
\caption{Hyper-parameter analysis of weights $\alpha$ and $\beta$ on M\textsuperscript{3}FD. Balanced weights produces the best performance.}
\centering
\renewcommand\arraystretch{1.1}
\setlength{\tabcolsep}{13pt}
\begin{tabular}{cc|ccc}
\toprule
$\alpha$  & $\beta$   & $mAP$     & $AP_{50}$  & $AP_{75}$\\
\hline
0.5×     &1.5×   & 42.0  &  68.3    &42.8 \\
1.5×     &0.5×   & 44.9  &  72.8    &45.6 \\
1.0×     &0.5×   & 45.3  &  74.6    &45.9 \\
0.5×     &1.0×   & 43.7  &  71.9    &44.1 \\
\cellcolor[HTML]{DAE8FC}1.0×     &\cellcolor[HTML]{DAE8FC}1.0×   &\cellcolor[HTML]{DAE8FC}\textbf{47.4}  &\cellcolor[HTML]{DAE8FC}\textbf{75.9}    &\cellcolor[HTML]{DAE8FC}\textbf{49.3} \\
\bottomrule
\end{tabular}
\label{tab:ablaion-balabce}
\end{table}

\begin{table}[t]
\caption{Analysis of convolution kernels with different numbers and sizes in CBR block on FLIR.}
\centering
\renewcommand\arraystretch{1.1}
\setlength{\tabcolsep}{9pt}
\begin{tabular}{l|ccc|c}
\toprule
Kernel      & Person     & Car  & Bicycle   &$AP_{50}$\\
\hline
-   &  90.1    & 89.1      & 72.3      & 83.8  \\
$\cellcolor[HTML]{DAE8FC}[1\times 1]$   & \cellcolor[HTML]{DAE8FC}\textbf{90.2}    &\cellcolor[HTML]{DAE8FC}\textbf{92.5}      & \cellcolor[HTML]{DAE8FC}\textbf{75.6}      &\cellcolor[HTML]{DAE8FC}\textbf{86.1}  \\
$[3\times3]$   &  88.6    & 92.0     & 73.9      & 84.8  \\
$[3\times3]\times2$   &  89.2    & 92.1      & 74.6      & 85.3  \\
\bottomrule
\end{tabular}
\label{tab:ablation-cbr-flir}
\vspace{-0.2cm}
\end{table}

\begin{table}[t]
\caption{Comparison of $AP$ on three benchmark datasets by utilizing our approach with the base detectors YOLOv7.
Our approach effectively improves detection accuracy across different datasets.}
\centering
\renewcommand\arraystretch{1.1}
\setlength{\tabcolsep}{2.5pt}
\begin{tabular}{c|c|cccccccc}
\toprule
Datasets                   & Method  & $mAP$    & $AP_{50}$  & $AP_{75}$ & $AP_S$   & $AP_M$   & $AP_L$  \\
\hline
\multirow{3}{*}{FLIR}  & baseline                & 44.2           & 84.2       & 39.5       & 32.3       & 49.3       & 64.3\\
                               &ours            & 45.8     & 86.1        & 41.1        & 34.6       & 50.7       & 66.8     \\
                               & \cellcolor[HTML]{DAE8FC}$\pm \Delta$                & \cellcolor[HTML]{DAE8FC}\textbf{+1.6}           & \cellcolor[HTML]{DAE8FC}\textbf{+1.9}        & \cellcolor[HTML]{DAE8FC}\textbf{+1.6}        & \cellcolor[HTML]{DAE8FC}\textbf{+2.3}       & \cellcolor[HTML]{DAE8FC}\textbf{+1.4}       & \cellcolor[HTML]{DAE8FC}\textbf{+2.5}      \\
\hline                         
\multirow{3}{*}{M\textsuperscript{3}FD}  & baseline                &47.6            & 78.3       & 49.0       & 19.0       & 59.1       & 79.7     \\
                                &ours            & 51.7     & 83.7        & 53.3        & 25.4       & 62.0       & 80.4     \\
                               & \cellcolor[HTML]{DAE8FC}$\pm \Delta$                & \cellcolor[HTML]{DAE8FC}\textbf{+4.1}           & \cellcolor[HTML]{DAE8FC}\textbf{+5.4}       & \cellcolor[HTML]{DAE8FC}\textbf{+4.3}       & \cellcolor[HTML]{DAE8FC}\textbf{+6.4}       & \cellcolor[HTML]{DAE8FC}\textbf{+2.8}       & \cellcolor[HTML]{DAE8FC}\textbf{+0.7}\\


\bottomrule
\end{tabular}
\label{tab:comparison-baseline}
\vspace{-0.3cm}
\end{table}

\begin{table}[t]
\caption{Analysis of convolution kernels with different numbers and sizes in CBR block on M\textsuperscript{3}FD.}
\centering
\renewcommand\arraystretch{1.1}
\setlength{\tabcolsep}{7pt}
\begin{tabular}{l|cccc|c}
\toprule
Kernel      & Bus     & Car  & Lamp      &Truck  & $AP_{50}$\\
\hline
-   & 89.4      & 91.2      & 70.8           & 87.4      & 83.2      \\
$\cellcolor[HTML]{DAE8FC}[1\times1]$   &\cellcolor[HTML]{DAE8FC}\textbf{88.5}      &\cellcolor[HTML]{DAE8FC}\textbf{90.9}      &\cellcolor[HTML]{DAE8FC}\textbf{71.7}        &\cellcolor[HTML]{DAE8FC}\textbf{87.4}      &\cellcolor[HTML]{DAE8FC}\textbf{83.7}      \\
$[3\times3]$   & 86.8      & 90.7      & 69.3          & 85.6      & 82.7      \\
$[3\times3]\times2$   & 87.3      & 90.3      & 70.2          & 86.8      & 83.1   \\
\bottomrule
\end{tabular}
\label{tab:ablation-cbr-m3fd}
\vspace{-0.2cm}
\end{table}

\subsection{Objects Quantity at Different FPN Layers}
We count the quantities of objects from each detection head. As illustrated in Figure~\ref{zhuzhuangtu} (a), the proportion of large-scale objects (P5 layer detects) in the two datasets is relatively small, accounting for about 7\% and 11\% of the total number of objects on the two datasets, respectively.
Consequently, we  ultimately choose to disentangle the P3 and P4 layers that correspond to detect small and medium-scale objects. And our method achieves a significant improvement of 1.92\% on the P3 and P4 layers.

\setcounter{figure}{0}
\begin{figure}[t]
    \centering
    \includegraphics[width=0.99\linewidth]{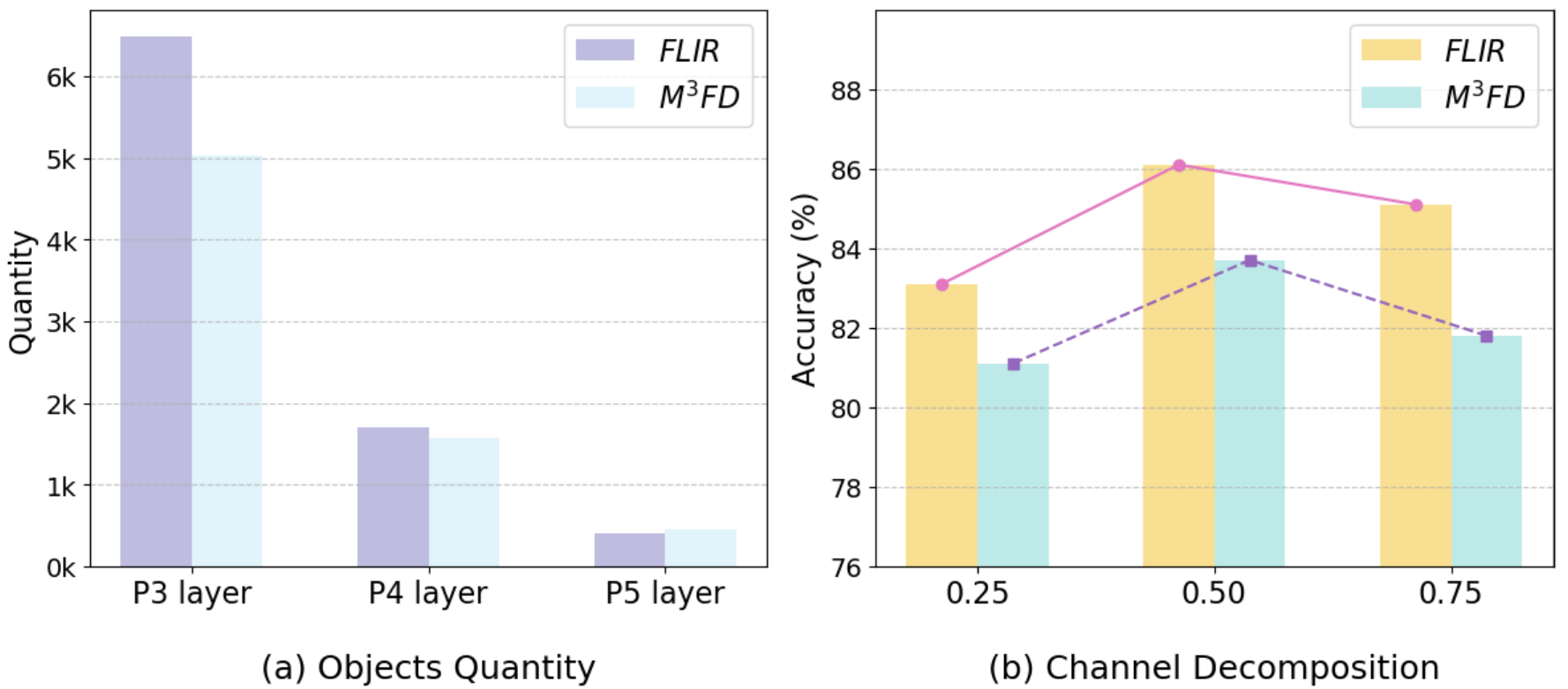}
    \caption{
    Statistical analysis of objects quantity and sensitivity analysis of channel decomposition. (a) The quantity of objects is calculated from different FPN layers on the two datasets. (b) The horizontal axis represents the proportion of decomposed object feature compared to the original feature.}
    \label{zhuzhuangtu}
\end{figure}

\subsection{Channel Decomposition Sensitivity Analysis}
The original features were divided into object features and non-object features. 
To investigate the impact of channel decomposition ratio on the final detection performance,
we conduct experiments with the ratio set at 0.25, 0.5, and 0.75, respectively.
The results are illustrated in Figure~\ref{zhuzhuangtu} (b). The best performance was achieved when the object features accounted for half of the original features. Accordingly, we adopted an even split of the original feature channels.

\subsection{Different Captions Generation Approaches}
\begin{figure}[!t]
    \centering
    \includegraphics[width=0.99\linewidth]{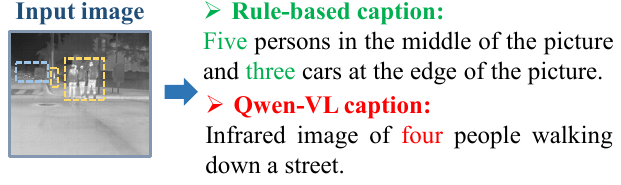}
    \caption{An example of natural-language description.}
    \label{caption_generate}
    \vspace{-0.2cm}
\end{figure}

We generate the natural-language descriptions of infrared images employing the Qwen-VL model. However, there are many noisy captions with Figure~\ref{caption_generate} as an example. 
This phenomenon is mainly attributed to the fact that the typical LVLM is not optimized for infrared modality and the hallucination problem tends to provide fake information. Furthermore, our approach still works with weakly supervised noisy captions.It can be extended to scenarios where detailed annotations are unavailable.

\section{Visual Analysis}

We visualized more object detection results on FLIR that we could not include in the main paper. 
From the visualization results in Figure~\ref{fig1}, the LGFD significantly disentangles the features into object and non-object components which split by  P3 and P4 layers of FPN. 
Therefore, we can conclude that the effectiveness of LGFD lies in its ability to concentrate on robust object features while suppressing background noise.

\begin{figure*}[t]
    \centering
    \includegraphics[width=0.99\linewidth]{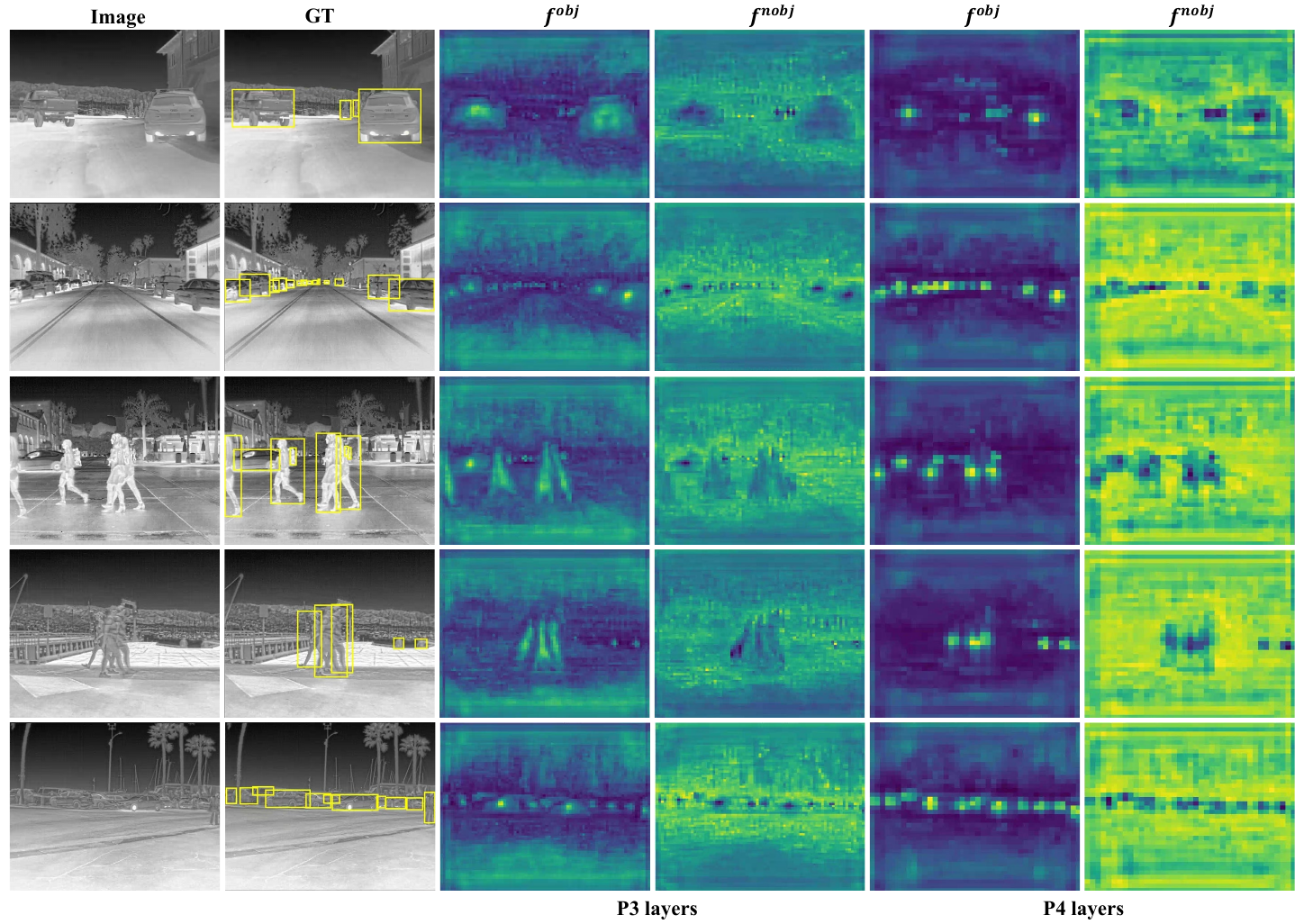}
    \caption{Visualization of object and non-object features of P3 and P4 layers in FPN. The object features are more focused on the foreground where the ground truth (GT) exists. Conversely, non-object features are primarily concerned with the background.}
    \label{fig1}
\end{figure*}

\end{document}